%% file: acl2023.tex
\pdfoutput=1

\documentclass[11pt]{article}

\input{macros}
\usepackage{svg}
\usepackage{enumitem}
\usepackage{graphicx}
\usepackage{multirow}
\usepackage{pifont}
\usepackage{arydshln}
\usepackage{subfigure}
\usepackage[final]{acl}

\finaltrue

\usepackage{times}
\usepackage{latexsym}

\usepackage[T1]{fontenc}

\usepackage[utf8]{inputenc}

\usepackage{microtype}

\usepackage{inconsolata}

\title{\texttt{M3T}: A New Benchmark Dataset for Multi-Modal Document-Level Machine Translation}

\author{Benjamin Hsu$^\Diamond$, Xiaoyu Liu\footnotemark[1]$^\dagger$, 
Huayang Li\footnotemark[1]$^\ddagger$,\\
\bf{Yoshinari Fujinuma$^\Diamond$,
Maria Nadejde$^\Diamond$,
Xing Niu$^\Diamond$}, \\
\bf{Yair Kittenplon$^\Diamond$,
Ron Litman$^\Diamond$,
Raghavendra Pappagari$^\Diamond$} \\
$^\dagger$ University of Maryland, College Park\hspace{5em}
$^\ddagger$Nara Institute of Science and Technology \\
  \hspace{2em}\texttt{xliu1231@umd.edu}\hspace{10em}
\hspace{1em} \texttt{li.huayang.lh6@is.naist.jp}\\
$^\Diamond$AWS AI Labs\\
\texttt{\{benhsu,fujinuy,mnnadejd,xingniu,yairk,litmanr,pappaga\}@amazon.com}
}

\begin{document}

\maketitle
\renewcommand*{\thefootnote}{\fnsymbol{footnote}}
\footnotetext[1]{Equal contribution. Work conducted during internships at Amazon.}
\renewcommand*{\thefootnote}{\arabic{footnote}}

\input{sections/0-abstract}

\input{sections/1-intro}
\input{sections/3-method}

\input{sections/4-experiments}
\input{sections/6-conclusion}
\input{sections/7-limitations-impact}
\input{sections/9-acknowledgements}

\bibliography{acl2023}
\newpage

\input{sections/8-appendix}

\end{document}

%% file: macros.tex
\newif\iffinal
\newcommand{\xn}[1]{\iffinal\else\textcolor{purple}{#1}\fi}
\newcommand{\bh}[1]{\iffinal\else\textcolor{magenta}{#1}\fi}

\newcommand{\yf}[1]{\iffinal\else\textcolor{cyan}{#1}\fi}

\newcommand{\dataset}{\texttt{M3T} }

%% file: sections/0-abstract.tex
\begin{abstract}



Document translation poses a challenge for Neural Machine Translation (NMT) systems. Most document-level NMT systems rely on meticulously curated sentence-level parallel data, assuming flawless extraction of text from documents along with their precise reading order. These systems also tend to disregard additional visual cues such as the document layout, deeming it irrelevant. However, real-world documents often possess intricate text layouts that defy these assumptions. Extracting information from Optical Character Recognition (OCR) or heuristic rules can result in errors, and the layout (e.g., paragraphs, headers) may convey relationships between distant sections of text. This complexity is particularly evident in widely used PDF documents, which represent information visually. This paper addresses this gap by introducing \dataset, a novel benchmark dataset tailored to evaluate NMT systems on the comprehensive task of translating semi-structured documents. This dataset aims to bridge the evaluation gap in document-level NMT systems, acknowledging the challenges posed by rich text layouts in real-world applications.

\end{abstract}

%% file: sections/1-intro.tex
\section{Introduction}

Traditional machine translation (MT) systems primarily focus on textual content at the sentence level, ignoring both global context and visual layout structure of a document.
Given that long-range contextual dependencies are crucial for generating high quality translations \cite{laubli-etal-2018-machine,  toral-etal-2018-attaining, hassan2018achieving}, these systems fail short of achieving human translation quality when considering entire documents \cite{junczys-dowmunt-2019-microsoft}. Several auhors have sought to address these gaps by including additional context, but these focused solely on text content (i.e. preceeding sentences) \citep{fernandes-etal-2021-measuring,lopes-etal-2020-document, tiedemann-scherrer-2017-neural}.
By treating documents solely as textual content, existing methods ignore a significant portion of the information encapsulated within the visual aspects of documents.

\begin{table}
  \begingroup
  \includegraphics[width=\columnwidth]{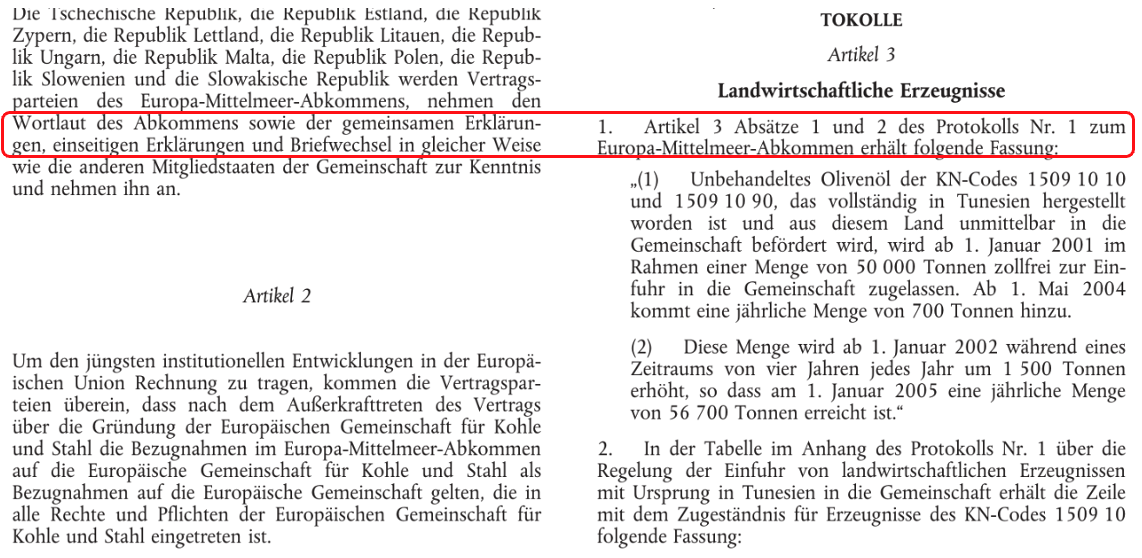}
  \captionof{figure}{Example OCR errors that occur in layout-agnostic systems. Reading order is lost when extracting text and column breaks are not correctly recognized.}\label{fig:ex_ocr}
  \endgroup

\vspace*{.5\floatsep}
  \centering
  \resizebox{\columnwidth}{!}{
    \begin{tabular}{ll}
    \multicolumn{2}{c}{OCR w/o layout}  \\ \hline
    Src 1 &  general principles for \textbf{implemet-kit} \\
    Src 2 & \textbf{ge comma vocstonal ele poly...} \\ \hline
    Tgt 1 &  allgemeiner Grundsätze für den Implementierungssatz \\
    Tgt 2 & ge Komma Gesang ele Poly... \\
    \hline \hline
    \multicolumn{2}{c}{} \\
    \multicolumn{2}{c}{OCR w/ layout}  \\ \hline
    Src & general principles for implementing  \\
    & a common vocational training policy... \\ \hline
    Tgt & allgemeiner Grundsätze für die Umsetzung  \\
    & einer gemeinsamen Berufsbildungspolitik... \\
    \hline \hline
    \end{tabular}}
  \caption{Without layout information, the downstream translation system cannot recover from OCR errors resulting in garbled translations. With layout information, the reading order and contiguous blocks of text are preserved, resulting in improved translation usability and quality.}
  \label{layout-aware:examples}
\end{table}

Visual cues represent an important yet overlooked set of features which can provide contextual clues. For instance, real world documents typically include different typeface in denoting section headings to different parts of the document which indicates to human readers that parts of a document should be treated as a logically distinct. Text blocks can also indicate that sections of text are contextually related and multi-columnar text is easily understood by human readers. Layout elements can also indicate that certain text should \textit{not} be translated (e.g. address or name fields) or that tabular data is present. Disregarding layout and visual information can lead to catastrophic translation errors, as demonstrated in Table \ref{layout-aware:examples}. 




\begin{figure}[!t]
    \centering
    \includesvg[width=\columnwidth]{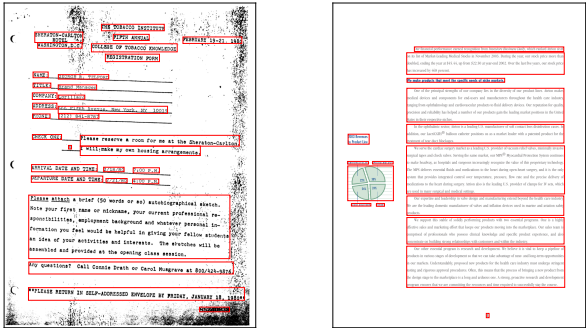}
    \caption{We introduce a benchmark dataset for PDF document translation. Unlike earlier works, our benchmark that focus on sentence-level text or cleanly segmented text, ours tests models on their ability to utilize visual queues for translating text. Above from left to right, are example documents sourced from the RVL-CDIP and DocLaynet datasets demonstrating the complex layout and domains included in the benchmark.  Additional examples can be found in the Appendix, Figure \ref{fig:example_annotations_app}.}
    \label{fig:example_annotations}
\end{figure}

In recent years, there has been significant focus on Vision-Language (VL) architectures, which has led to notable advancements in multi-modal reasoning \citep{li2022blip, alayrac2022flamingo,ganz2023towards, li2023blip, ganz2024question}. Specifically, recent research has dedicated considerable attention to understanding structured documents, particularly in the domain of intelligent document processing and information extraction (e.g. visual question answering) tasks \citep{Xu2020LayoutLMPO, xu2020layoutlmv2,huang2022layoutlmv3, tang-etal-2022-udop}. In terms of translation, earlier work has explored correcting single-word translation using visual features \cite{nikolov-etal-2018-character, salesky-etal-2021-robust, wang2020word, biten2022latr}.
With recent advances in multi-modal foundation models (FMs) \citep{alayrac2022flamingo, openai2023gpt4, liu2023visual, tang-etal-2022-udop}, that combine visual encoders with large language models (LLMs), development of multi-modal models to handle visually and textually challenging tasks like translation is within reach. The ability to benchmark model performance on a challenging document understanding task that takes into account long range contextual clues is of growing importance.

Publicly available datasets for benchmarking document-level MT mainly consist of plain text documents with sentence or paragraph level alignment. The widely used JRC-Acquis consists of 8,000 documents in 20 official EU languages aligned at sentence and paragraph level \citep{steinberger-etal-2006-jrc}. Similarly, the Web Fiction benchmark \citep{wang-EtAl:2023:WMT} consists of manually aligned sentence pairs that capture book and chapter-level contexts. These benchmarks address the text domain, assuming perfect segmentation of document structure (e.g. document and paragraph boundaries).

In this work, we present a multi-modal benchmark dataset for evaluating MT models on translating visually rich PDF documents. Our dataset is unique in that it focuses on machine translation at the document level and test models on both their ability to translate and their ability to use visual features as contextual clues. Our contributions are: (1) We introduce \dataset, a new challenging benchmark testset for evaluating end-to-end \textbf{M}ulti-\textbf{M}odal \textbf{MT} of structured documents (Figure~\ref{fig:example_annotations}). To further aid in research in document understanding, we have also annotated layout and reading order of the extracted text; (2) We conduct initial experiments using a multi-modal foundation model, LLaVa-v1.5 \citep{liu2023visual}, and find that multi-modal features improve translation quality, though there is  significant room for improvement; (3) Finally, we release synthetically generated parallel data to aid future model development by the community.\footnote{Dataset and scripts can be found here: \url{http://github.com/amazon-science/m3t-multi-modal-translation-bench}} 



%% file: sections/3-method.tex
\section{Dataset Description}

\begin{table}[!th]
    \centering
    \resizebox{\columnwidth}{!}{
    \begin{tabular}{l| ccc}
       	& RVL-CDIP & DocLayNet	& EUR-Lex \\
       	\hline\hline
    \multicolumn{4}{l}{Layout Elements}  \\
       	\hline
    Text &	2780 & 6679 & 1325 \\
    Section-header & 144 &	1070 &	999 \\
    Page-footer	& 92 & 685 & 0 \\
    Vertical	& 68 & 133 & 1088 \\
    Table-cell	& 1332	& 16292 & 16451 \\
    Table	    & 44 & 256	& 445\\
    Page-header	& 220 &	525 & 1532 \\
    Picture	& 28 & 444	& 0 \\
    List-item &	204 & 2406 & 1317 \\
    Formula	& 0 & 23 &	2 \\
    Footnote &	0 &	93 & 391 \\
    Code	& 0	& 13 & 0 \\
    Title	& 0	& 74 & 30 \\
    Caption	& 0	& 76 & 24 \\
    \hline \hline
    \multicolumn{4}{l}{Language Pairs}  \\
       	\hline
    en$\rightarrow$de & 20 & 126 & 0 \\
    de$\rightarrow$en  &  0 & 34 & 54 \\
    en$\rightarrow$es & 20 & 126 & 0 \\
    es$\rightarrow$en & 0 & 0 & 54 \\
    en$\rightarrow$fr & 20 & 126 & 0 \\
    fr$\rightarrow$en &	 0 & 11 & 54 \\
    en$\rightarrow$zh & 20 & 126 & 113 \\
    zh$\rightarrow$en &	 0 & 19 &  0 \\
    \hline\hline
    \multicolumn{4}{l}{Document Domains}  \\
       	\hline
    Scanned &  80 & -	 & - \\
    Govt. tenders	& - &102 & - \\
    Patents	& - & 36 & - \\
    Legal & - &	71 & 383 \\
    Fin. reports	& - & 344 & - \\
    \hline
    \end{tabular}
    }
    \caption{Distribution of the layout elements, language pairs, and document domains contained in the benchmark dataset. 
    Language pairs indicate the original language of the source document.
   }
    \label{dataset:sampling}
\end{table}

\dataset focuses on PDF documents which are a commonly utilized format that pose several challenges for modern language models. Even in digitally generated PDFs, certain artifacts, such as white characters strategically placed for spacing adjustments or the duplication of characters in the absence of bold-face options, can complicate straightforward text extraction.

\paragraph{Data Sourcing}

We sourced documents from several public datasets to cover a wide range of documents. First, we used EUR-Lex\footnote{\url{https://eur-lex.europa.eu/homepage.html}} to source documents which are translated by professional translators into all the European languages while also preserving the document layout. We sample a subset of documents which we annotated with layout information described in the next section. 
\xn{What does "additionally translated" mean? We re-use translations provided in EUR-Lex? (If no, why?) Tables/captions/footnotes are not translated in EUR-Lex?}

Second, we sourced documents from DocLayNet~\cite{doclaynet} and RVL-CDIP~\citep{rvl_cdip}, which consist of documents annotated with layout information. We sourced annotations on the documents from their respective test sets to preserve the train/test split that might be used in training layout/vision models (e.g. LayoutLM). 

For the sampling strategy, we focused on covering a wide range of layout elements. For RVL-CDIP, were randomly selected, excluding fax cover sheets. For DocLayNet we sampled documents containing flat layouts, multi-columnar data, images and tabular information. For EUR-Lex, we used an initial set of automatically labeled layout elements for sampling documents, and selected documents containing more than just text elements (e.g. tables, captions, footnotes) that were not previously translated and represent more complex layout features. Initial automatic labelling was done using a Faster R-CNN classifier on the DocLayNet dataset using the standard \texttt{dectrctron2}\footnote{https://github.com/facebookresearch/detectron2} following \citet{doclaynet}. Our classifier had an F1 of 73.4 averaged across all layout elements. 

\paragraph{Document Annotation} 

We annotated the sampled documents in two stages: (1) Annotators first extracted the text and labeled the layout elements; (2) Annotators translated the text extracted in each layout element. Table \ref{dataset:sampling} reports the final distribution of layouts and domain coverage.

For the layout information, annotators were asked to provide bounding box coordinates and labels for each element. Bounding box labeling means grouping the elements on the page into reasonable paragraphs, headings, tables, pictures, etc. areas on the page, in non-overlapping rectangular areas (see Figure \ref{fig:example_annotations}). For the set of labels, we extended the DocLayNet ontology to include reading order and table cells. Specifically, annotators labelled the layout elements described in Table \ref{annotation:labels}.

For translating the extracted text, we provided professional translators with the original document and asked to take into account the context of the document when translating. Extracted text was then translated using a selection of commercial segment level MT systems. Translators were then instructed to post-edit these translations with the additional requirement that the length of the translation within $\pm$10\% of the source text length, which is useful for evaluating how well systems may preserve the original layout of the source document. In order to facilitate length compliance, we sourced documents in de, es, and fr and asked to translate them into en. We also asked to translate en documents to into zh. Additional annotation details can be found in Appendix \ref{sec:annoation_details}.

\paragraph{Supplementary Data}

We also provide automatically annotated EUR-Lex documents for future research (e.g., model training). First, we downloaded the plain text parallel corpus based on EUR-Lex \citep[JRC-Acquis]{steinberger-etal-2006-jrc} and extracted text from EUR-Lex documents using Tesseract OCR.\footnote{\url{https://github.com/tesseract-ocr/tesseract}} Next, we aligned text in JRC-Acquis with recognized (OCR'd) bounding boxes in EUR-Lex documents. Finally, we erased unaligned bounding boxes and produced tuples of \texttt{<source-text, OCR'd-text, bounding-box, target-text>}.

%% file: sections/4-experiments.tex
\section{Multi-Modal Document Translation}

\begin{table}[t]
    \centering
    \resizebox{\columnwidth}{!}{
    \begin{tabular}{l|lll|c|c|}
     & \multicolumn{3}{c|}{Visual Features} &  \multicolumn{2}{c|}{doc-COMET}   \\
    \hline
    Text & Rand. & Block & Doc & EUR-Lex & DocLayNet\\
    \hline
    \multirow{3}{.5cm}{OCR} 
    & \ding{51} & & & 0.0173 & 0.0120 \\
    & & & \ding{51} & \bf{0.0174} & \bf{0.0126} \\
    & & \ding{51}  & & 0.0174 & 0.0122\\
    \hline
    \multirow{3}{1cm}{Gold} 
    & \ding{51} & & & 0.0214 & 0.0193 \\
    &  & & \ding{51} & \bf{0.0218} & \bf{0.0194} \\
    &  & \ding{51} & & 0.0215 & 0.0192\\
    \hline
    \end{tabular}}
    \caption{Results from our experiments using LLaVa-v1.5 using different visual features such as random images (Rand.), images of text blocks (Block), and images of the whole document (Doc). We tested how these features are used to improve OCR'd text (OCR) and if they provide additional context for clean text (Gold). We found that visual features improved document level translation quality in terms of doc-COMET scores. Results are averaged over de, fr$\rightarrow$en language pairs. We found LLaVa is capable of using visual features to address OCR errors. It also appears to generate better translations when given the document image as context. }
    \label{table:results}
\end{table}

Multi-modal document translation is under explored because of a lack of annotated benchmark datasets. Given the creation of a novel benchmark dataset in this work, we conducted experiments testing how recent multi-modal FMs perform on multi-modal document translations.
To motivate future research, we experiment with a recent LLaVa-v1.5 model \cite{liu2023visual}, a large multi-modal model that combines a CLIP vision encoder ViT-L~\cite{pmlr-v139-radford21a} and Vicuna-13b \cite{zheng2023judging} for general-purpose visual and language understanding. 

We conduct experiments adding various visual and layout features to the model. For visually rich documents, evaluating OCR text translation directly is difficult since an alignment between  extracted source segments with clean source segments needs to be found. We leave the development of such an alignment model to future work. Instead, our experiments focus on whether the visual features are utilized by a multi-modal FM to improve translations by (1) recovering from OCR errors or (2) understanding the context of the document. We focused our experiments on X$\rightarrow$en documents for reasons discussed below.

To test these hypothesizes, we controlled for the effects of incorrect layout prediction
by conducting experiments using the gold annotated layout elements (e.g., annotated bounding boxes). For hypothesis (1), we experimented with different granularity of visual features (at the text block level or at the document level). We used Tesseract for OCR in our experiments to extract text blocks and passed a single text block at a time to the LLaVA model. We also compared with using translations gold source text translation to understand the upper bound on translation quality. Finally to test hypothesis (2), we experimented with translations using the gold text to see if less granular visual features enable a multi-modal FM to understand contextual clues better. As a baseline, we tested random images from the MS COCO dataset~\citep{mscoco} into the LLaVa model which allows us to measure improvements from including relevant visual features. We conducted ablations using images of identified text blocks or images of the entire document. Lastly for evaluation, since the reading order of the text was annotated, we used document level COMET scores\footnote{https://github.com/amazon-science/doc-mt-metrics} \cite{vernikos-etal-2022-embarrassingly} with a context size of two (i.e., two previous segments as context).

\paragraph{Results}

\yf{We likely need an ablation study on injecting synthetic OCR errors to further support this claim.} \bh{+1}
Table~\ref{table:results} summarizes the results from our experiments. 
We found that visual features helped the model improve translations of OCR'd text. We found evidence for this comparing performance on different languages. Tesseract OCR performance is known to vary significantly for none Latin script languages \citep{gupte2021genalog}. In particular, on Chinese text, we found that including images in the model improved performance by +0.02 when including images of the entire document.

\xn{Do we measure OCR errors?} We found that including relevant visual features improved upon random visual features from the MS COCO dataset. Moreover, features focused on the relevant text blocks improved translation quality (as measured by doc-COMET score) on both EUR-Lex and DocLayNet subsets of data. When looking at document level visual features, we found that these universally improved translation quality beyond the baseline with random visual features.

While there were improvements, visual features did not help translation appreciably. This presents an opportunity for future researchers to find methods to leverage contextual clues in the visual domain more fully. One issue was that the Vicuna-13b model suffered from hallucinations especially when the inputs are short, single word, or numerical segments which is the case for many of the tabular data from both EUR-Lex and DocLayNet documents. In the case of en$\rightarrow$X translation, we found that the model misinterpreted the prompt and did not produce translations.

%% file: sections/6-conclusion.tex
\section{Conclusions}

In conclusion, we release \dataset, a new benchmark for assessing multi-modal machine translation of visually-rich documents. We conducted experiments using a multi-modal FM built ontop of Vicuna-13b and found that while these models attain impressive results on visual question-answering and captioning tasks, multi-modal document translation is still an area for future research. 

%% file: sections/7-limitations-impact.tex
\section{Limitations}

In this work, we used available open source models which were not specifically tuned for this benchmark task. Finally, while we sourced documents from large set of domains, the benchmark is ultimately limited to a few cases and heavily focused on legal and financial text in a few high resource languages (de, es, fr, and zh) which can introduce bias in future model evaluation and development.
 Furthermore, we did not experiment extensively with prompts, and an interesting future direction could be an investigation into whether their are better prompts to use for multi-modal models. 

%% file: sections/9-acknowledgements.tex
\section*{Acknowledgements}

We thank Tanya Badeka and Jen Wang for their help in curating the dataset and Yogarshi Vyas and Stanislas Lauly for their comments and suggestions early in this project.

%% file: sections/8-appendix.tex
\appendix
\onecolumn
\section{Annotation Details}
\label{sec:annoation_details}
For layout annotations, professional annotators were instructed to annotate bounding box coordinates and assign a label to the type of text given the definitions outlined in Table \ref{annotation:labels}. A random selection 10\% of annotation per annotators was selected and manually reviewed. If more than 5\% of those labels did not follow guidelines, the set was then re-annotated.

For translations, we followed a similar protocol. We used multiple systems to generate initial translations for each sentence per document at the segment level.  Professional translators were then asked to post-edit translations based on the context of the document. A 10\% of translations were sampled and checked for errors. If there were errors, these samples were re-translated.

\begin{table*}
\resizebox{.95\textwidth}{!}{
    \begin{tabular}{l|p{0.88\textwidth}}
    Layout & Description \\ \hline
    \textbf{Captions} & Special text outside a picture or table that describes the picture or table. \\ \hline
    \textbf{Footnotes} & Typically small text at the bottom of a page, with a number or symbol that is referred to in the text above. \\ \hline
    \textbf{Formula}& Mathematical equation or chemical formula on its own line. \\ \hline
    \textbf{List-item}& One element of a list, in a hanging shape, i.e., from the second line onwards the paragraph is indented more than the first line. \\ \hline
  \textbf{Page-footer}& Repeating elements like page number at the bottom, outside of the normal text flow. \\ \hline
    \textbf{Page-header}& Repeating elements like page number at the top, outside of the normal text flow. \\ \hline
    \textbf{Picture}& A graphic or photograph. \\ \hline
    \textbf{Section-header}& Any kind of heading in the text, except overall document title. \\ \hline
    \textbf{Table}& Material arranged in a grid alignment with rows and columns, often with separator lines \\ \hline
    \textbf{Table-cell}& Text contained in a cell of a table. \\ \hline
    \textbf{Text-block}& Regular text paragraphs or blocks of text that should be logically grouped together. \\ \hline
    \textbf{Title}& Overall title of a document typically appearing in large font or denoted with special typeface to distinguish it from the rest of the document. \\ \hline
    \textbf{Order}& The order number that the bounding box should be read on the page (i.e. the reading order). For European languages, this is generally left to right, top to bottom. For tables, we adopt a left to right, top to bottom ordering of the cells. \\
    \hline
    \textbf{Vertical} & Text is oriented vertically.\\
    \hline
    \end{tabular}}
    \caption{Layout  elements and their labels annotated in our benchmark following the DocLayNet typology.}
    \label{fig:example_annotations_app} 
\end{table*}

\section{Detailed Results}
\begin{table*}[h]
    \centering
    \resizebox{.85\textwidth}{!}{
    \begin{tabular}{l|lll|ccc|ccc|}
     & \multicolumn{3}{c|}{Visual Features} &  \multicolumn{6}{c|}{doc-COMET}   \\
    \hline
     &  &  & & \multicolumn{3}{c|}{EUR-Lex} & \multicolumn{3}{c|}{DocLayNet}\\
    \cline{5-10}
      Text & Rand &  Block &  Doc & de$\rightarrow$en & fr$\rightarrow$en & es$\rightarrow$en & de$\rightarrow$en & fr$\rightarrow$en & zh$\rightarrow$en \\
    \hline
    \multirow{3}{.5cm}{OCR} 
    & \ding{51} & & & 0.0159  & 0.0187  & 0.0226 & 0.0123  & 0.0118  & -0.011 \\
    & & & \ding{51} & 0.0160  & 0.0187 	& 0.0225 & 0.0125  & 0.0128  & 0.013 \\
    & & \ding{51}  & & 0.0159 & 0.0189 	& 0.0226 & 0.0123  & 0.0120  & -0.010 \\
    \hline
    \multirow{3}{1cm}{Gold} 
    & \ding{51} & & & 0.0190 & 0.0238 & 0.0281 & 0.0219 & 0.0166 &  0.013 \\
    &  & & \ding{51} & 0.0195 & 0.0241 & 0.0282 & 0.0217 & 0.0170 & 0.014 \\
    &  & \ding{51} & & 0.0191 & 0.0238 & 0.0281 & 0.0219 & 0.0166 & 0.014\\
    \hline
    \end{tabular}}
    \caption{Results from our experiments using LLaVa-v1.5 using different visual features such as random images (Rand.), images of text blocks (Block), and images of the whole document (Doc). We tested how these features are used to improve OCR'd text (OCR) and if they provide additional context for clean text (Gold). We found that visual features improved document level translation quality in terms of doc-COMET scores. Results are averaged over de, fr$\rightarrow$en language pairs. We found LLaVa is capable of using visual features to address OCR errors. It also appears to generate better translations when given the document image as context. }
    \label{table:results_app}
\end{table*}

\onecolumn
\begin{figure}[ht]
    \centering
    \subfigure[]{\includesvg[width=.85\textwidth]{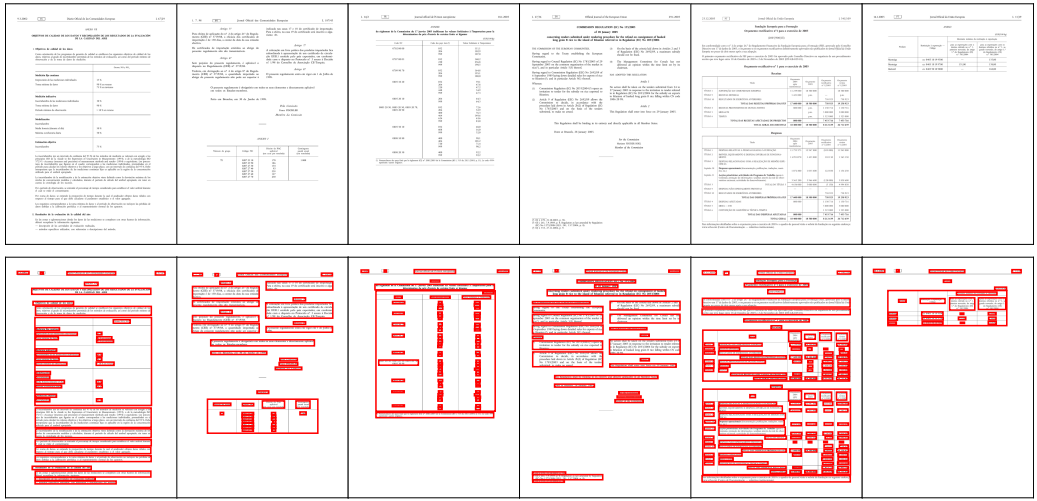}} 
    \subfigure[]{\includesvg[width=.85\textwidth]{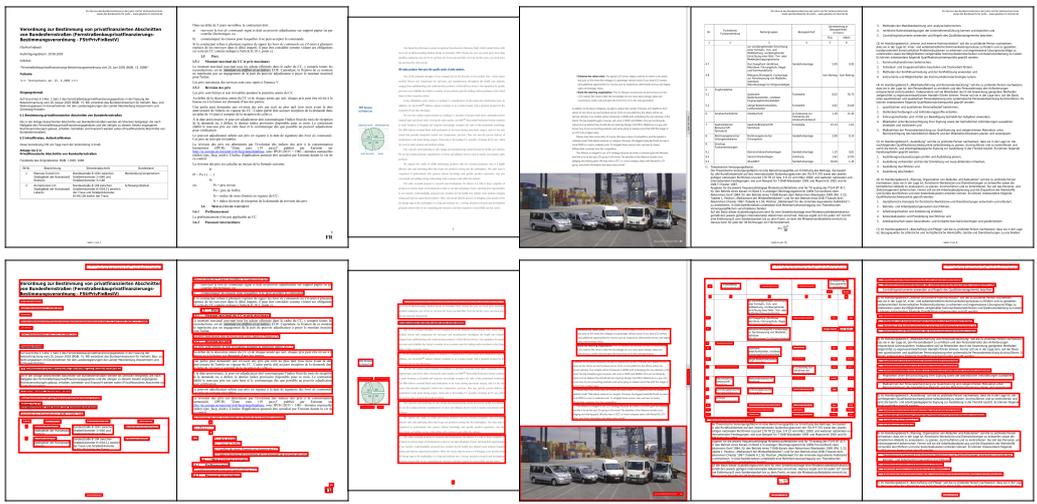}} 
    \subfigure[]{\includesvg[width=.85\textwidth]{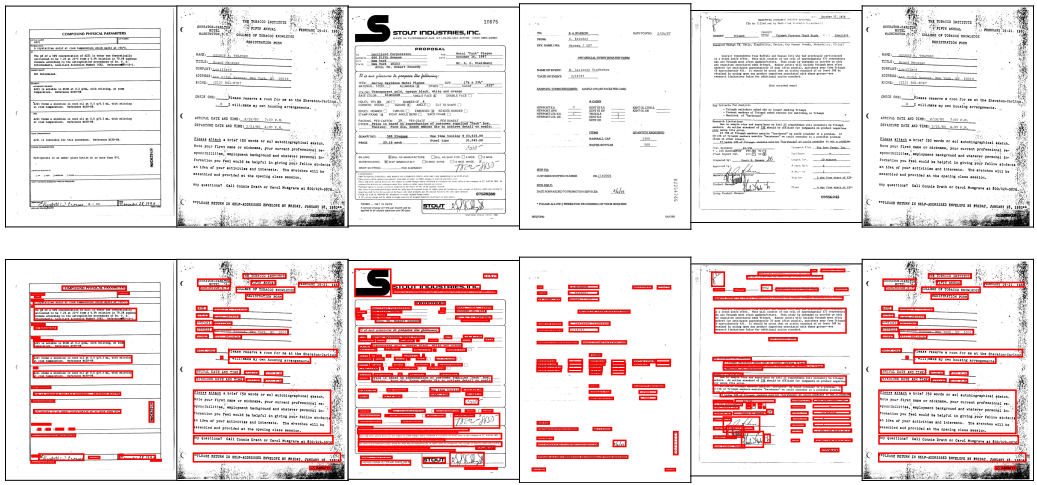}}
    \caption{In this work, we introduce a benchmark dataset for PDF document translation. Example documents sourced from (a) EUR-Lex (b) DocLayNet and (c) RVL-CDIP datasets demonstrating the complex layout and domains included in the benchmark. Unlike earlier works, our benchmark focuses on machine translation which depends more heavily on contextual information than earlier works centered around entity extraction.}
\end{figure}